\documentclass[sigconf]{acmart}

\copyrightyear{2022}
\acmYear{2022}
\setcopyright{acmcopyright}\acmConference[MM '22]{Proceedings of the 30th ACM International Conference on Multimedia}{October 10--14, 2022}{Lisboa, Portugal}
\acmBooktitle{Proceedings of the 30th ACM International Conference on Multimedia (MM '22), October 10--14, 2022, Lisboa, Portugal}
\acmPrice{15.00}
\acmDOI{10.1145/3503161.3548422}
\acmISBN{978-1-4503-9203-7/22/10}

\AtBeginDocument{%
  \providecommand\BibTeX{{%
    \normalfont B\kern-0.5em{\scshape i\kern-0.25em b}\kern-0.8em\TeX}}}
\definecolor{amber}{rgb}{0.98, 0.93, 0.36}
\usepackage{multirow}
\usepackage{enumitem}
\usepackage{makecell} 
\usepackage{longtable}

\newcommand{\figref}[1]{Figure~\ref{fig:#1}}

\newcommand{\tabref}[1]{Table~\ref{tab:#1}}
\newcommand{\tatdqa}{\textsc{TAT-DQA}}
\newcommand{\tatqa}{\textsc{TAT-QA}}
\newcommand{\tagop}{\textsc{TagOp}}
\newcommand{\model}{\textsc{MHST}}

\newcommand{\roberta}{RoBERTa$_\text{LARGE}$}

\newcommand{\llmvt}{LayoutLMv2$_\text{LARGE}$}
\newcommand\fone{F\textsubscript{1}}

\acmSubmissionID{mmfp3171}

\begin{document}

% \title{Discrete Reasoning Over Visual-rich Financial Documents}
\title{Towards Complex Document Understanding By Discrete Reasoning}


\author{Fengbin Zhu$^{1,2}$, Wenqiang Lei$^{3\ast}$, Fuli Feng$^{4}$, Chao Wang$^{2}$, Haozhou Zhang$^{3}$, Tat-Seng Chua$^1$}

\affiliation{%
  \institution{$^1$National University of Singapore, Singapore}
  \country{}
}
\affiliation{%
  \institution{$^2$6Estates Pte Ltd, Singapore} 
  \country{}
}
\affiliation{%
  \institution{$^3$Sichuan University, China} 
  \country{}
}
\affiliation{%
  \institution{$^4$University of Science and Technology of China, China} 
  \country{}
}

\email{{zhfengbin, wenqianglei, fulifeng93}@gmail.com, wangchao@6estates.com, dcscts@nus.edu.sg}

\thanks{$^\ast$Corresponding author}

\renewcommand{\shortauthors}{Zhu, Fengbin, et al.}

\begin{abstract}
Document Visual Question Answering (VQA) aims to answer questions over visually-rich documents.
In this work, we introduce a new Document VQA dataset, named \tatdqa, which consists of 3,067 document pages comprising semi-structured table(s) and unstructured text as well as 16,558 question-answer pairs.
The documents are sampled from financial reports and contain lots of numbers, which means discrete reasoning capability is demanded to answer the questions.
Based on \tatdqa, we further develop a novel model named \model~that takes into account the information in multi-modalities to intelligently address different types of questions with corresponding strategies, i.e., extraction or reasoning. 
The experiments show that \model~model significantly outperforms the baseline methods, demonstrating its effectiveness.
However, the performance still lags far behind that of expert humans.
We expect that our \tatdqa~dataset would facilitate the research on understanding of visually-rich documents, especially for scenarios that require discrete reasoning.
Also, we hope the proposed model would inspire researchers to design more advanced Document VQA models in future. 
Our dataset will be publicly available for non-commercial use at~\url{https://nextplusplus.github.io/TAT-DQA/}.

\end{abstract}

%     <concept>
%         <concept_id>10002951.10003317.10003347.10003348</concept_id>
%         <concept_desc>Information systems~Question answering</concept_desc>
%         <concept_significance>500</concept_significance>
%     </concept>
%   <concept>
%       <concept_id>10002951.10003227.10003251.10003253</concept_id>
%       <concept_desc>Information systems~Multimedia databases</concept_desc>
%       <concept_significance>500</concept_significance>
%       </concept>
%     <concept>
%         <concept_id>10010147.10010178.10010179</concept_id>
%         <concept_desc>Computing methodologies~Natural language processing</concept_desc>
%         <concept_significance>100</concept_significance>
%     </concept>
%     <concept>
%         <concept_id>10010147.10010178.10010179.10003352</concept_id>
%             <concept_desc>Computing methodologies~Information extraction</concept_desc>
%             <concept_significance>100</concept_significance>
%     </concept>

% \ccsdesc[500]{Information systems~Multimedia databases}
% \ccsdesc[100]{Computing methodologies~Natural language processing}
% \ccsdesc[100]{Computing methodologies~Information extraction}


\maketitle  

\section{Introduction}

Document understanding and analysis are indispensable in businesses of diverse domains like finance, legal, medical, etc~\cite{subramani2020survey}.
Such work is mostly performed manually, which is labor-intensive and time-consuming with low scalability~\cite{cui2021docai}. 
Intelligent Document Understanding (IDU) emerges as an important research area in multimedia, which spans Natural Language Processing (NLP) and Computer Vision (CV)~\cite{cui2021docai}. 
It aims to automatically read and understand business documents.
Many IDU tasks have been proposed, including Document Layout Analysis \cite{zhong2019publaynet,li2020docbank}, Table Detection and Recognition~\cite{zhong2019image,li2020tablebank,smock2021pubtables1m}, Key Information Extraction (KIE)~\cite{, Huang2019SROIE,Jaume2019FUNSD,Gralinski2020Kleister}, Document Visual Question Answering (VQA)~\cite{Mathew2020DocVQA,VisualMRC2021,mathew2021infographicvqa}, etc.

\begin{figure}[!t]
    % \toprule
    \begin{center}
    \includegraphics[scale=0.105]{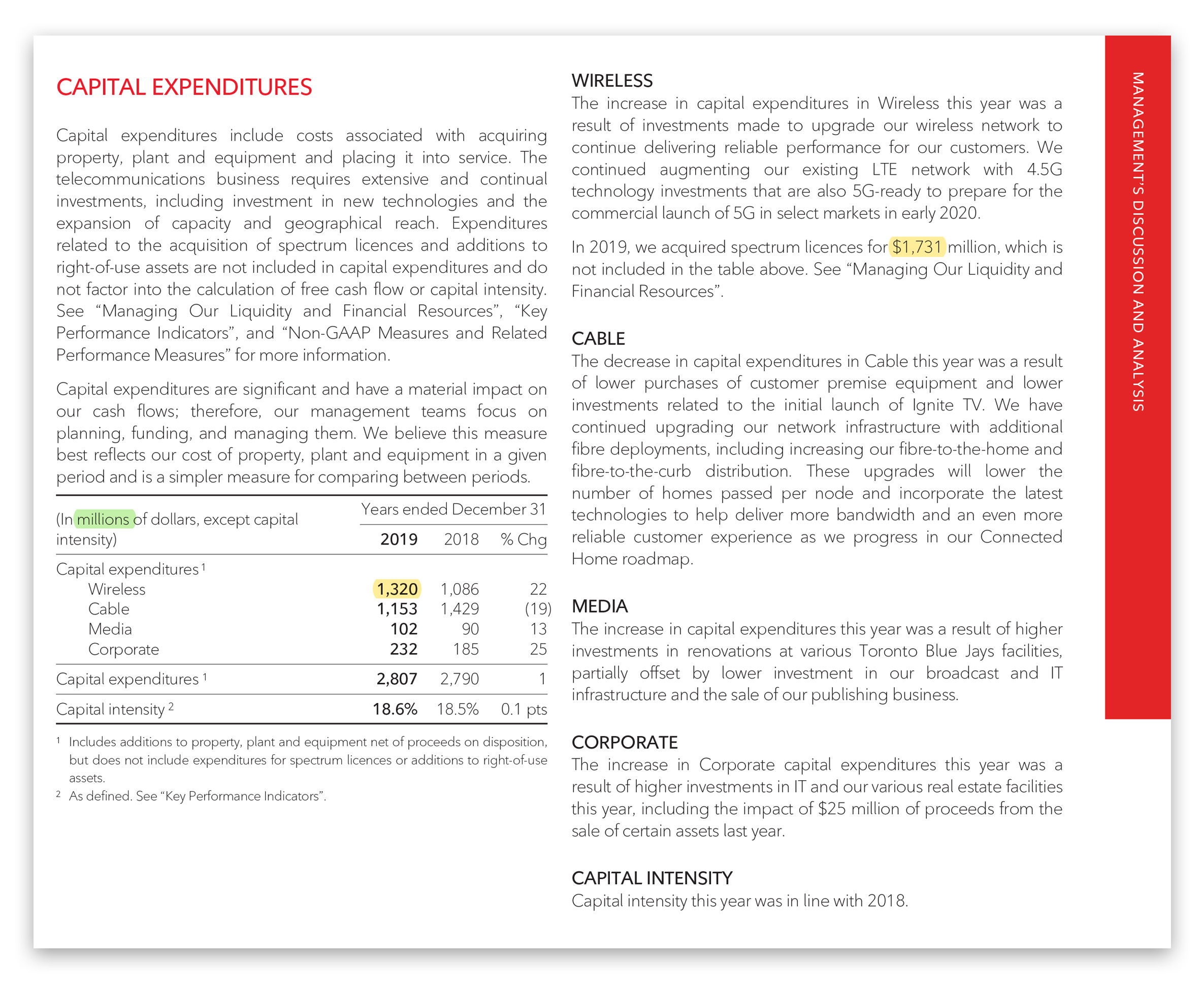}
    \end{center}
    \vspace{-1em}
    \scalebox{1.0}{
    \begin{tabular}{p{26em}}
        \textbf{Question}: What was the total cost in Wireless including spectrum license fee in 2019? \\
        \textbf{Derivation}: \colorbox{amber}{1,320} + \colorbox{amber}{1,731}  = 3,051\\
        % \textbf{Calculated}: 3,051 \\ 
        \textbf{Scale}: \colorbox{green}{Millions}\\
        \textbf{Answer}: 3,051,000,000 \\ 
    \end{tabular} }
    %   \toprule
    \caption{\label{fig:sample} 
    An example of \tatdqa~ dataset. 
    Given a question and a visually-rich document that contains both tabular and textual data, the machine is expected to derive the answer.
   }
\end{figure}

Among these tasks, Document VQA is a high-level document understanding task wherein given a visually-rich document and a relevant question in natural language (\figref{sample}), the model is required to give the correct answer to the question based on the document~\cite{cui2021docai}.
In Document VQA, the model needs to effectively exploit and harness the textual and layout information of the document besides its image information, compared to traditional VQA tasks~\cite{Yash2017vqa2,singh2019textvqa,biten2019scene} where image information is the focus.
For this task, we are particularly interested in handling those documents with semi-structured tables that usually contain numbers in addition to unstructured text.
This document type is informative and very pervasive in real-world businesses, however with only a few prior efforts~\cite{zhu2021tat, chen2021finqa} on auto-understanding them.
To facilitate the research on these problems, we construct a new \textbf{D}ocument VQA dataset by extending the \textbf{\tatqa}~\cite{zhu2021tat}, called \textbf{\tatdqa~} dataset.

The documents in \tatdqa~dataset are sampled from real-world high-quality financial reports and each document contains both tabular and textual data.
Furthermore, these documents contain lots of numbers, which means discrete reasoning capability, such as counting, sorting, comparison, addition, subtraction, multiplication, division and the compositions of them, is demanded to answer questions.
The average length of the documents in \tatdqa~is significantly larger than all existing Document VQA datasets.
To our best knowledge, \tatdqa~ is the first Document VQA dataset that is constructed based on real-world high-quality business documents.
Also, this work is the first one that attempts to understand the documents with multiple pages in literature on document understanding, which is more challenging compared to the understanding tasks over single-page documents.

Based on \tatdqa, we further propose a novel multi-modal Document VQA model, named \textbf{\model}.
To represent the question and document, the \model~ employs a multi-modal Transformer encoder to take the question as well as the document text, layout and visual image information as input. 
After that, to infer the answer, it first adopts a ``\textbf{M}ulti-\textbf{H}ead'' classifier to predict the answer type, i.e., \emph{Span}, \emph{Spans}, \emph{Counting} and \emph{Arithmetic}, based on which different prediction strategies are applied.
For the answer type of \emph{Span}, a Span Answer Predictor~\cite{rajpurkar2016squad,Devlin2018Bert} is applied to predict the starting and ending positions of the answer span; for the other three answer types, a Sequence Tagging Module is applied to extract the evidences for deriving the answer.
After obtaining the evidences, an Expression Tree Generator following ``\textbf{S}eq2\textbf{T}ree'' architecture is used to generate an expression tree to infer the final answers for arithmetical questions; for \emph{Spans} and \emph{Counting} questions the final predictions are obtained by collecting or counting all non-contiguous spans in the evidences.
Experiments show that \model~model significantly outperforms baseline methods, demonstrating effectiveness.
% However, its performance still lags far behind that of expert humans.

We expect that our \tatdqa~dataset would facilitate future research on the deep understanding of complex real-world documents combining vision and language, especially for scenarios requiring discrete reasoning, and that our \model~ model would inspire the community to develop more advanced Document VQA models.

\section{Related Work}
In this section we briefly review  previous research in Intelligent Document Understanding, the datasets for Document VQA, as well as discrete reasoning, with special attention to those works that are most related to ours.  

\subsection{Intelligent Document Understanding}
Intelligent Document Understanding (IDU) is to enable a machine to automatically read, understand, and analyze business documents.
This research area is very practically meaningful and much demanded, which spans both the Natural Language Processing (NLP) and the Computer Vision (CV) fields~\cite{cui2021docai}.
Thanks to the wide success of Transformer~\cite{Ashish2017attention,Devlin2018Bert} in addressing NLP and CV problems, Transformer-based models have been popularly applied to solving document understanding tasks.
The documents mainly involve three modalities of information: text, layout, and visual information.
Some works~\cite{xu2020layoutlm,li2021structurallm,hong2021bros,Garncarek2020LAMBERT} combine document layout information with text information in the Transformer encoder.
LayoutLM~\cite{xu2020layoutlm} and StructuralLM~\cite{li2021structurallm} incorporate document layout information as new positional embeddings into the embedding layer.
BROS~\cite{hong2021bros} and LAMBERT~\cite{Garncarek2020LAMBERT} take into account the spatial distance between tokens when computing the attention weights and bias in the self-attention layers.
Recently, some works~\cite{xu2021layoutlmv2,Appalaraju2021ICCV,Wu2021LAMPRET} propose to model all three modalities of information, which has become a defacto approach for almost all document understanding tasks.
% For example, LayoutLM v2~\cite{xu2021layoutlmv2} obtains visual embeddings from the document image using a CNN-based visual encoder~\cite{xie2017Aggregated} and appends them in the input sequence;
% DocFormer~\cite{Appalaraju2021ICCV} proposes a discrete multi-modal Transformer encoder architecture and fuses text and vision features in each self-attention layer with the help of spatial information.
In this work, we also adopt a Transformer-based model to encode document text, layout and visual image information for our task.

\subsection{Document VQA Datasets}
Document VQA is a high-level document understanding task wherein a model is required to answer a question in natural language given a visually-rich document~\cite{cui2021docai}.
To date, there are only a few datasets that have been proposed specially for Document VQA, to our best knowledge, including DocVQA~\cite{Mathew2020DocVQA}, VisualMRC~\cite{VisualMRC2021}, InfographicVQA~\cite{mathew2021infographicvqa} and WebSRC~\cite{chen2021websrc}.
Among them, InfographicVQA focuses on infographic instead of documents, and WebSRC does not provide its statistics in the original paper.
DocVQA is constructed using various types of industry documents for extractive question answering, where answers can always be extracted from the text in the document.
In DocVQA, the documents are mostly from the 1960-2000 period, with low-resolution, and besides some high-quality documents with printed or digital-born text, there are also some with typewritten and handwritten text.
The VisualMRC~\cite{VisualMRC2021} dataset is built for abstractive question answering, where answers cannot be directly extracted from the text in the document. 
The documents in VisualMRC and WebSRC are screenshots of web pages.
In this work, we build a new and complex \tatdqa~ dataset with real-world high-quality business documents, hoping to facilitate future research in the community.

\subsection{Discrete Reasoning}
Discrete reasoning is key to solving many NLP tasks, like Math Word Problems (MWPs)~\citep{kushman2014learning,wang2017deep,huang2017learning,xie2019goal,zhang2020graphtree} and Question Answering (QA) over text~\citep{Dua2019DROP,Zhu2021Retrieving}, tables~\citep{Pasupat2015Compositional} and both~\citep{zhu2021tat,chen2021finqa,li2022learning}.
In a Math Word Problem, it is required to answer a mathematical query according to a textual description and has been studied since the 1960s~\cite{bobrow1964natural}. 
The MVPs are often solved by generating an expression tree to derive the final answer implicitly~\cite{wang2018translating,chen2019semantically} or explicitly~\cite{liu2019tree,xie2019goal,zhang2020graphtree}.
For discrete reasoning in textual QA like the DROP~\cite{Dua2019DROP} dataset, researchers usually employ a ``Multi-Head'' classifier to predict the answer type first and then perform arithmetic operations to derive the final answer  \cite{hu2019multi,segal2020simple,ran2019numnet,andor2019giving,chen2020question}.
For discrete reasoning over tables only or a combination of the table and text, recent works incorporate the table structure feature in the positional encoding layer~\cite{herzig2020tapas,wang2021tuta}, or the attention layer~\cite{yin2020tabert,wang2021tuta} to jointly train tables and text.
To our best knowledge, no prior work attempts to develop models that are capable of performing discrete reasoning over real-world business documents. 

\section{Proposed Dataset: TAT-DQA}
In this section, we present the definition of the Document VQA task, the construction of the \tatdqa~dataset, and the statistical analysis of the dataset. 

\subsection{Task Definition}

Consider a visually-rich document $D$ of one or more pages, which contains both tabular and text contents.
Each page is converted to an image plus a list of words using the PDF/OCR converter.  
Given a question $Q$, the model $\mathcal{F}$ is required to predict the answer $a$ according to the document $D$ as demonstrated in \figref{sample}.
Formally, the task is formulated as
\begin{align}
    \mathcal{F}(D, Q) &= a. 
\end{align}

In \tatdqa, the answer value $a$ may be either extracted from the given document, or generated by performing discrete reasoning such as addition, subtraction, multiplication, division, counting, comparison/sorting, and their compositions.

To solve this task, a multi-modal Document VQA model usually needs to take into account the document text content, layout information, and visual image information in order to derive the final answer.
In this process, the capability of discrete reasoning over the visually-rich document is much demanded.

\begin{table}[t]
\centering
\begin{tabular}{@{}lrrr@{}}
\toprule
{\bf Statistic} & {\bf Train} & {\bf Dev} & {\bf Test}  \\ 
\midrule
    \# of questions & 13,251 & 1,645 & 1,662 \\
    \# of documents & 2,207 & 274 & 277 \\
    \# of documents (1 page) & 2,004 & 229 & 219 \\
    \# of documents (>1 page) & 203 & 45 & 58 \\
    Avg. length of words (per document) & 539.90 &	579.89 &603.57 \\
    Avg. length of words (per page) & 494.43 & 496.53	& 496.42 \\
    \bottomrule
    \end{tabular}
\caption{Basic statistics of each split in \tatdqa}
\vspace{-1em}
    \label{tab:tatdqa_stats}    
\end{table}

\subsection{Dataset Construction}
The dataset \tatdqa~ is built upon a previous \tatqa~\cite{zhu2021tat} dataset. 
In \tatqa, each \textit{hybrid context} (i.e. data) comprises a well-structured table and some relevant texts. They are all selected and sorted manually by human experts from financial reports in PDF format. With human interference, each hybrid context contains only one table and the corresponding texts are sure to semantically correlate with the table.
Compared with \tatqa, the \tatdqa~ better aligns with real scenarios. 
Each document in \tatdqa~may contain one or more tables, and the texts inside this document may correlate with the table(s), or may have nothing to do with the table(s). 
This task setting is much challenging than that on \tatqa.

\subsubsection{Collection of Question-Answer Pairs}
\label{sec-qa}
To construct the new \tatdqa~ dataset, we borrow the question-answer (Q-A) pairs from the previous \tatqa~ dataset, which are generated by human experts in finance. On this basis, we ask human experts in finance to generate some more Q-A pairs, and meanwhile remove a few pairs with errors we have found during data preparation.
In total, we get $16,558$ Q-A pairs for \tatdqa.
The same as \tatqa, \tatdqa~ offers four types of answers: 
\begin{itemize}
\item \emph{Span}: The answer is a continuous text in the document~\cite{rajpurkar2016squad};
\item  \emph{Spans}: This type of answer is also called ``Multi-span" and is a set of non-contiguous spans in the document~\cite{Dua2019DROP}; 
\item \emph{Counting}: The answer is an integer that is computed by performing counting;
\item \emph{Arithmetic}: The answer is a numerical value that is obtained by performing arithmetic operations such as addition, subtraction, multiplication, division and their compositions.
\end{itemize}

Also, the scale of the answer in \tatdqa~can have five values, including \texttt{None}, \texttt{Thousands}, \texttt{Millions}, \texttt{Billions}, \texttt{Percent}.
To facilitate development of discrete reasoning capability of Document VQA models, we also keep the corresponding derivations for \emph{Counting} and \emph{Arithmetic} questions in the same formats with \tatqa. 

\subsubsection{Collection of Document Pages}
To construct the new \tatdqa~dataset, we filter out and keep the document pages corresponding to the above acquired Q-A pairs from the raw financial reports.
These raw financial reports are real-world ones, mostly dated between 2018 and 2020, which are downloaded from the web\footnote{https://www.annualreports.com/} based on the file names provided by the authors of \tatqa.
Each Q-A pair corresponds to one document, and a document may contain at most three pages.
We process each retaining document page after filtering to obtain the text with a bounding box by using Apache PDFBox\footnote{https://pdfbox.apache.org/} for the readable PDF files or a commercial OCR engine for the images.
Finally, each document page is converted to a list of text blocks and each text block has a list of words, with every block or word framed by its own bounding box.
In total, we obtain $2,758$ documents consisting of $3,067$ pages. 
The document itself and its text content with the corresponding bounding box will be released together in \tatdqa.

\begin{table}[t]
\centering
\vspace{-1em}
\begin{tabular}{lrrrr}
\toprule
\bf Answer Type & \bf Train & \bf Dev & \bf Test & \bf Total \\
\midrule
Span & 5,737 & 690 & 714 & 7,141 \\
Spans & 1,656 & 216 & 210 & 2,082 \\
Counting & 305 & 32 & 40 & 377 \\
Arithmetic & 5,553 & 706 & 699 & 6,958 \\
Total & 13,251 & 1,645 & 1,662 & 16,558 \\
\bottomrule
\end{tabular}
\caption{Question distribution regarding different answer types of each split in \tatdqa.}
\vspace{-1em}
    \label{tab:tatdqa_qtype}  
\end{table}

\subsection{Statistics and Analysis}
In total, the \tatdqa~dataset includes $2,758$ documents, consisting of $3,067$ document pages from $182$ financial reports, and $16,558$ question-answer pairs.
These documents are randomly split into a training set ($80\%$), a development set ($10\%$) and a test set ($10\%$); all the questions about a particular document belong to only one of the splits.
We summarize the basic statistics of each split in \tabref{tatdqa_stats}, and the question distribution regarding the answer type in \tabref{tatdqa_qtype}.

A comparison of our new \tatdqa~ dataset with the two existing document VQA datasets DocVQA~\cite{Mathew2020DocVQA} and VisualMRC~\cite{VisualMRC2021} is summarized in \tabref{tatdqa_comparison}.
In particular, for \textbf{document type}, the documents in \tatdqa~stem  from real-world high-quality financial reports between 2018 and 2020, and each document must contain both tabular and textual data. 
Comparably, the documents of DocVQA are from the UCSF Industry Documents Library, which are mostly within the 1960-2000 period, of low-resolution and include some typewritten and handwritten text; 
VisualMRC is built with the screenshots of web pages instead of real-world documents.
For \textbf{document length}, the average length of documents in \tatdqa~(550.29 words) is significantly larger than that of DocVQA (182.75 words) and VisualMRC (151.46 words), which makes \tatdqa~more complex and challenging.
For the \textbf{number of pages per document},
each document in DocVQA or VisualMRC has only 1 page; in contrast, \tatdqa~has at most 3 pages in the document, and its average page number per document is 1.33 and over 11.0\% documents in \tatdqa~have more than 1 page.
We also compare the \textbf{answer type} of the three datasets: \tatdqa~consists of extractive (i.e., \emph{Span} and \emph{Spans}) and abstractive (i.e., \emph{Counting} and \emph{Arithmetic}) answers that need to be generated with discrete reasoning;
DocVQA only has SQuAD-like extractive and short answers while VisualMRC focuses on long abstractive answers.

\begin{table}[t]
\centering
\small
\begin{tabular}{l|r|r|r}
\toprule
\bf Property & \bf DocVQA & \bf VisualMRC & \bf \tatdqa \\
\midrule
Document Type & \begin{tabular}{l} Industry \\ document\end{tabular}  & \begin{tabular}{l} Web \\ pages\end{tabular}  & \begin{tabular}{l} Finance \\ reports\end{tabular}   \\
\midrule
Document Period & 1960 - 2000 & Jan - Mar 2020 & 2018 - 2020 \\
\midrule
Avg len of document & 182.75 & 151.46 & 550.29 \\
Avg no. of pages & 1 & 1 & 1.33 \\

\midrule
Answer Type & Ext. & Abs. &  Ext. +  Abs. \\
\bottomrule
\end{tabular}
\caption{The comparison among the three Document VQA datasets, i.e., DocVQA, VisualMRC and \tatdqa. The length of the document is counted in terms of OCR words. \emph{Ext}. and \emph{Abs} denote \emph{extractive} and \emph{abstractive} respectively.}
    \label{tab:tatdqa_comparison}  
\end{table}

To the best of our knowledge, the \tatdqa~dataset is the first document VQA dataset that is constructed on top of  real-world high-quality business documents.
It is also the most complex document VQA dataset till now, with more than one page per document.
We promise to release this dataset in near future, hoping to facilitate the research on document understanding techniques in the community. 

%%%%%%%%
\section{Proposed Method: \model}

\begin{figure*}[t]
    \begin{center}
    \includegraphics[scale=0.9]{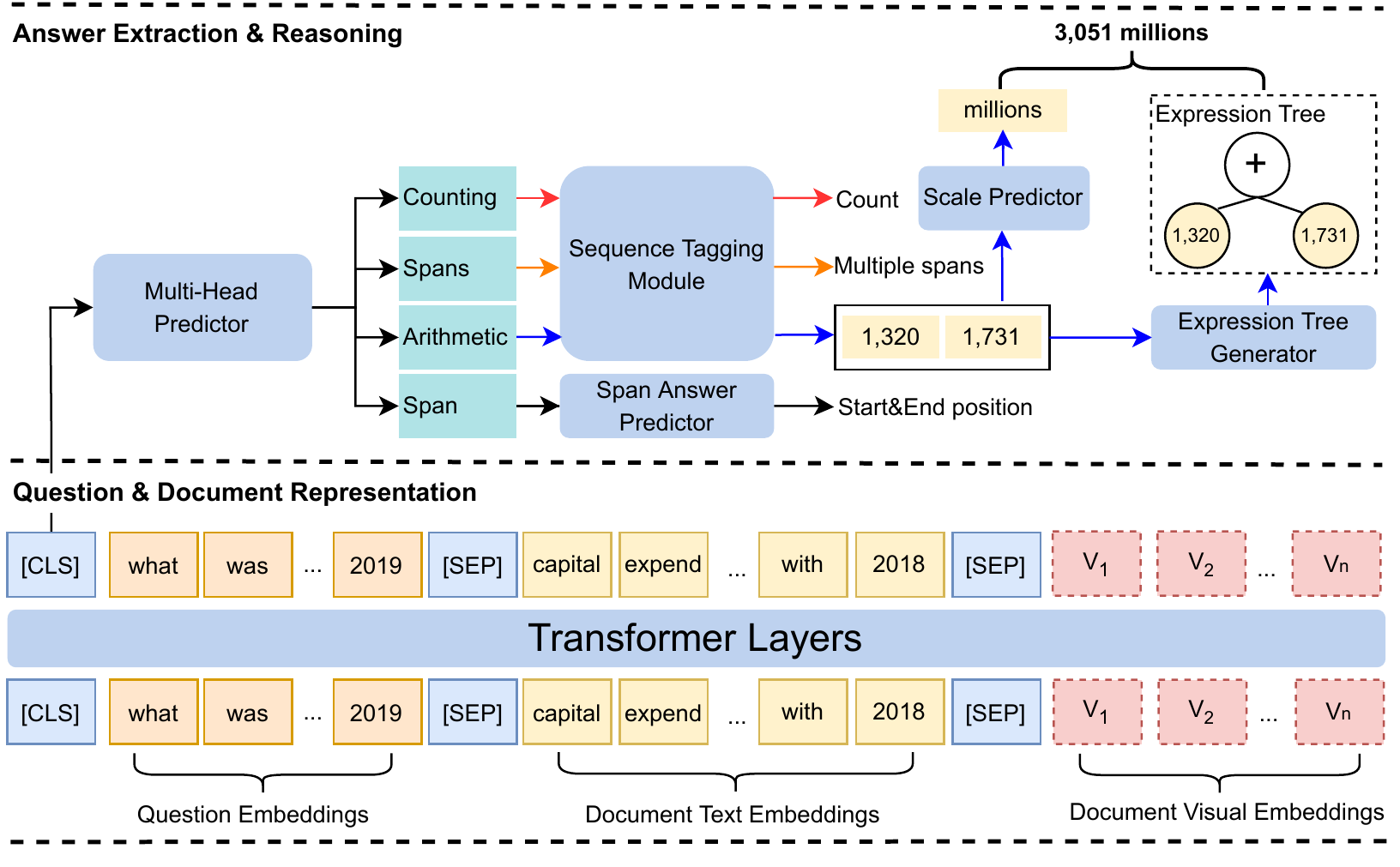}
    \end{center}
    \vspace{-1em}
    \caption{\label{fig:model}
    The overall architecture of the model~\model.
    Take the question in \figref{sample} as an example. \model~employs a multi-modal Transformer architecture to take the text, layout and visual embeddings as input.
    The visual embeddings (i.e., V1, V2,...,Vn) are obtained by leveraging a CNN-based visual encoder.
    Then it adopts a ``\textbf{M}ulti-\textbf{H}ead'' classifier to predict the answer type, i.e., \emph{Span}, \emph{Spans}, \emph{Counting} and \emph{Arithmetic}.
     An Expression Tree Generator following ``\textbf{S}eq2\textbf{T}ree'' architecture is applied to generate an expression tree with the selected numbers by Sequence Tagging Module to derive the final answers for  the arithmetical questions.
     }
\end{figure*}

The data type in the new \tatdqa~ dataset is very common in real-world business scenarios, i.e. a document of one or more pages containing one or more tables as well as several texts.
To effectively auto-understand such documents, we introduce a novel multi-modal QA model, named \model, which has two stages, i.e., \textbf{Question and Document Representation}, and \textbf{Answer Extraction and Reasoning}.
To represent the question and the document, the \model~employs a multi-modal Transformer architecture to encode the question, document text, layout and visual image in the input.
To infer the final answer, the \model~adopts a ``\textbf{M}ulti-\textbf{H}ead'' classifier to predict the answer type, i.e., \emph{Span}, \emph{Spans}, \emph{Counting} and \emph{Arithmetic}.
For the answer type of \emph{Span}, a Span Answer Predictor~\cite{rajpurkar2016squad,Devlin2018Bert} is applied to predict the starting and ending positions of the answer span; for the other three answer types, a Sequence Tagging Module is applied to extract the evidences for deriving the answer.
After obtaining the evidences, an Expression Tree Generator following ``\textbf{S}eq2\textbf{T}ree'' architecture is applied to generate an expression tree to infer the final answers for arithmetical questions; for \emph{Spans} and \emph{Counting} questions the final predictions are obtained by collecting or counting all non-contiguous spans in the evidences.
An overall architecture of the \model~model is illustrated in \figref{model}.

\subsection{Question and Document Representation}
\label{subsec4-1}
The \model~model adopts \llmvt~\cite{xu2021layoutlmv2} to generate the representations of the question and the document, which is a recent popular multi-modal Transformer model for document understanding. 
The \model~model takes as input a question as well as the text content, layout, and visual image information of the document.

The document text and layout information in the input can be obtained as follows. 
Each document in \tatdqa~contains both tabular and textual data.
We empirically find that the performance would be better if we handle them separately, as is verified in the experiment of Section \ref{exp-challenge}. 
To identify the table(s) vs. the text in the document, we propose to apply a heuristic method to identify the text blocks that belong to the table in each document page.
Then we apply TF-IDF to sort the rest non-table text blocks by estimating the similarity score of each block with respect to the question, considering that the average length of the document in \tatdqa~ is quite long, as demonstrated in \tabref{tatdqa_stats}.

To obtain the visual information of the document for the input, we transform the document page\footnote{We select the document page that has the major table for the multi-page documents.} in PDF to an image, which is resized to 224 x 224 and then fed into the image encoder proposed in~\cite{xu2021layoutlmv2} to obtain the visual embedding.
Finally, the embeddings of the question, table blocks,  non-table blocks and the image are input sequentially to the \llmvt~model to obtain question token representations $\overline{Q}$ and document token representations $\overline{D}$.

\subsection{Answer Extraction and Reasoning}
After obtaining the representations of the question and document,  \model~further identifies the answer type of the question and applies the corresponding strategies to derive the final answer.
\subsubsection{Multi-Head Predictor}
We design a Multi-Head Predictor to predict the answer type, i.e., \emph{Span}, \emph{Spans}, \emph{Counting} and \emph{Arithmetic}, as explained in Section \ref{sec-qa}.
This Multi-Head Predictor is essentially a multi-class classifier.
In particular, we take the vector of \texttt{[CLS]} as input to compute the probability of each answer type:
\begin{align}
    \mathbf{p}^\textrm{head} &= \textrm{softmax}(\textrm{FFN}(\texttt{[CLS]}))
\end{align}
where FFN denotes a two-layer feed-forward network with the GELU activation~\cite{Dan2016GELU}. 
For the answer type of \emph{Span}, a Span Answer Predictor is applied to predict the starting and ending positions of the answer span following the typical SQuAD-like QA~\cite{rajpurkar2016squad,Devlin2018Bert}.
For the other three answer types, we apply a Sequence Tagging Module to extract the corresponding evidences of the answer inspired by~\cite{zhu2021tat}, but use Beginning-Inside–Outside (\texttt{BIO}) tagging~\cite{segal2020simple} instead of Inside–Outside (\texttt{IO}) tagging to better handle multi-span answers.
After obtaining the evidences, the final prediction for \emph{Spans} and \emph{Counting} questions are obtained by collecting or counting all non-contiguous spans in the evidences following \tagop~\cite{zhu2021tat}.

\subsubsection{Expression Tree Generator}
For the question whose answer type is predicted as \emph{Arithmetic}, we apply an Expression Tree Generator to generate an expression tree to compute the answer~\cite{wang2018translating,xie2019goal,zhang2020graphtree}.
The Expression Tree Generator in our \model~ is implemented with the Goal-driven Tree Structure (GTS)~\cite{xie2019goal}.
GTS is a tree structured neural network that generates expression trees in a goal-driven manner, demonstrating noticeable effectiveness in solving Mathematical Word Problems (MWPs).
However, a typical MWP involves only a handful of numbers,  while in  \tatdqa, one document usually contains much more numbers, which significantly overwhelms the capacity of GTS.
To address this issue, we propose to select several most relevant numbers with the Sequence Tagging Module, and feed them into the GTS.

The expression trees generated by the Generator contain three kinds of nodes: the arithmetical operators $V_{op}$ (i.e., +-*/), the constant numbers $V_{con}$, and the numbers $V_{tag}$ that are identified by the Sequence Tagging Module in the document $D$, which form the target vocabulary $V^{dec}$ of the document $D$.
The constant numbers $V_{con}$ and the numbers $V_{tag}$ selected from the document are always set to be in leaf nodes positions.
The operators $V_{op}$ will always occupy the non-leaf nodes positions, and each operator node must have two child nodes.
The construction of the tree starts from producing the topmost operator, followed by the left child node, which will be repeated until the leaf node is produced. 
Then, the right child nodes are generated recursively. 
As such, the Generator generates an equation following the pre-order traversal ordering.

To start tree generation process of GTS, our model initializes the topmost root node vector using the vector of \texttt{[CLS]}.
For each token \texttt{t} in target vocabulary $V^{dec}$, its embedding $\mathbf{e}(t|D)$ is defined as
\begin{equation}
\label{eq2}
    \mathbf{e}(t|D) =
    \begin{cases}
    \mathbf{M}_{op} &  \text{if t} \in V_{op} \\
    \mathbf{M}_{con} &   \text{if t}  \in V_{con} \\
    \mathbf{h}_{loc}(t,D) &  \text{if t} \in V_{tag}.
    \end{cases}
\end{equation}
The representations of the numbers in $V_{tag}$ are document-dependent;
i.e., they will take the corresponding $h_{loc}(t, D)$ from $\overline{D}$.
However, the representations of the operators and the constant numbers are independent of the document, which   are obtained by two independent embedding matrices $\mathbf{M}_{op}$ and $\mathbf{M}_{con}$ respectively.

\subsubsection{Scale Predictor}
If the answer type is \emph{Arithmetic}, the scale predictor takes as input the concatenation of the vector of the $[CLS]$ token and the mean of the representations of all predicted numerical values to compute the probability of each scale:
\begin{align}
    \mathbf{p}^\textrm{scale} &= \textrm{softmax}(\textrm{FFN}([\texttt{CLS},h^\textrm{N}]))
\end{align}
where FFN denotes a two-layer feed-forward network with GELU activation and $h^\textrm{N}$ is obtained by computing the mean of all representations of the predicted numerical tokens by the Sequence Tagging Module.
For the types of \emph{Span}, \emph{Spans} and \emph{Counting}, we adopt the same method in \tagop, which takes the vector of $[CLS]$ only as input.
After obtaining the scale, the final answer is derived by multiplying or concatenating the answer value with the scale, depending on whether the answer value is a number or a string.

\subsection{Training and Inference}

To train our model, the objective is to minimize the negative log-likelihood loss which is the sum of the losses of all above classification tasks depending on the answer types, i.e., Multi-Head Predictor, Sequence Tagging Module, Span Answer Predictor, Scale Predictor and Expression Tree Generator.
Note that the ground-truths of the sequence tagging predictions are extracted from the annotated answers and derivations. 
For the arithmetic type, if the numerical values that are used to generate the ground-truth expression tree are not predicted, we will also add them in the ground truth in order to train the Expression Tree Generator.

During inference, our model first chooses the answer type and then performs the corresponding prediction strategies.
For the \emph{Span} type, the span with the maximum probability is attained as the final prediction among all the valid spans.
If the answer type is \emph{Spans} or \emph{Counting}, we collect or count all the predicted non-contiguous spans as the final prediction.
Following ~\cite{xie2019goal}, we apply a beam search to select the best expression tree and execute it to obtain the final prediction for the arithmetical questions.

\section{Experiments}
In this section, we report and analyze the extensive experimental results to demonstrate the validity of our new \tatdqa~dataset and also the promising effectiveness of our proposed Document VQA model \model~by comparing it with two baseline models. 

\subsection{Baseline Models}
To the best of our knowledge, there are very limited models that have been proposed to effectively solve QA tasks over the documents containing both tabular data and textual data, where discrete reasoning is particularly demanded.   
We choose two state-of-the-art QA models that have demonstrated promising discrete reasoning capability for comparison, i.e. NumNet+ V2~\cite{ran2019numnet} and \tagop~\cite{zhu2021tat}.

NumNet+ V2~\cite{ran2019numnet} is a textual QA model with impressive performance on DROP~\cite{Dua2019DROP} dataset that requires discrete reasoning over the textual data.
It constructs a numerically-aware graph neural network, which takes all the numbers in the given question and passage as nodes and builds edges via numerical comparison, and performs discrete reasoning over the graph to infer the final answer.
To adapt the model to \tatdqa, we apply TF-IDF between the question and each text block to sort the text blocks and convert them to a sequence as the input to the model, which is similar with the method introduced in Section~\ref{subsec4-1}.

The other model, \tagop~\cite{zhu2021tat}, achieves the state-of-the-art results on the \tatqa~dataset that requires discrete reasoning over a well-structured table and its relevant text to derive the answer.
It first employs a sequence tagging module to identify relevant cells from the table and spans from the text, and then applies a set of aggregation operators (e.g., addition, subtraction, multiplication, division, counting,etc) over them to infer the final answer.
To enable the model to work on the new \tatdqa~dataset, we omit the processing of the table, apply TF-IDF to sort the text blocks in the document and feed them to the model sequentially.

\begin{table}[t]
    \centering
    \begin{tabular}{lrrrr}
    \toprule
     \multirow{2}{*}{\bf Method}    & \multicolumn{2}{c}{\bf Dev} & \multicolumn{2}{c}{\bf Test} \\
     \cmidrule(lr){2-3}
     \cmidrule(lr){4-5}
         & EM & \fone{} & EM&  \fone{}\\
    \midrule
    \bf Human Experts  &  - & - & 84.1 & 90.8\\
    
    \addlinespace
    \multicolumn{5}{l}{\bf Baselines }\\
    NumNet+ V2   &  28.1 & 36.6 & 30.6 & 40.1\\
    \tagop   &  32.3 & 39.6 & 33.7 & 42.5 \\
    \addlinespace

    \multicolumn{5}{l}{\bf Text Only}\\
    \model~(\roberta) &  37.1 & 43.6 & 39.8 & 47.6 \\
    \addlinespace

    \multicolumn{5}{l}{\bf Text + Image}\\
    \model~(\llmvt) &  \bf 39.1 & \bf 47.4 & \bf 41.5 & \bf 50.7 \\

    \bottomrule
    \end{tabular}
    \caption{Performance of our model and baseline models on dev and test set of \tatdqa. Best results are marked in bold.}
   
    \label{tab:tatdqa-metric}
\end{table}

\subsection{Evaluation Metrics}
We adopt the same evaluation metrics used in~\cite{zhu2021tat} to measure the model performance on the \tatdqa~ dataset, i.e. the Exact Match (EM) and the numeracy-focused (macro-averaged) \fone{} score, taking into account the scale and the plus-minus of a numerical value.
Both Exact-Match and numeracy-focused (macro-averaged) \fone{} score measure the overlap between a bag-of-words representation of the gold and predicted answers.
Note that the numeracy-focused \fone{} score is set to 0 unless the predicted number multiplied by the predicted scale equals exactly the ground truth.

\subsection{Results and Analysis}
In experiments, we first compare our \model~model with two baseline models, NumNet+ V2 and \tagop, by testing their performance on understanding the documents in \tatdqa~dataset via Document VQA tasks.
We then analyze effects of our \model~model over each answer type in \tatdqa.  
Also, we experiment on \tatqa~dataset~\cite{zhu2021tat} to further verify the effectiveness of our proposed model.
Finally, based on all experimental results, we analyze the challenges of our new \tatdqa~dataset to reveal its properties. 

\subsubsection{Comparison with Baselines on \tatdqa}
The experimental results of our proposed model \model~and the baseline models are shown in \tabref{tatdqa-metric}.
We train two different variants of our \model~model by using different modalities in \textbf{Question and Document Representation} stage.
The first variant is the full model \model (\llmvt),  which adopts \llmvt~as the encoder and takes the  question as well as document text, layout, and visual image as input.
The other one is \model~(\roberta), which employs \roberta~as the encoder instead to represent the document with textual features only.
From \tabref{tatdqa-metric}, we can observe that both the two variants significantly outperform the baseline methods, demonstrating the effectiveness of our \model.
The superior performance of our full model \model (\llmvt) highlights the importance of the fusion of vision and language features to successfully answering the questions in \tatdqa.
But, there is still a big gap compared to the human performance.  

\begin{table}[t]
    \centering
    \begin{tabular}{lrrrr}
    \toprule
     \multirow{2}{*}{\bf Answer Type}    & \multicolumn{2}{c}{\bf Dev} & \multicolumn{2}{c}{\bf Test} \\
     \cmidrule(lr){2-3}
     \cmidrule(lr){4-5}
         & EM & \fone{} & EM&  \fone{}\\
    \midrule

  Span & 39.0 & 53.5 & 41.1 & 58.3 \\
  Spans & 26.9 & 43.0 & 25.7 & 43.3 \\
  Counting & 46.3 & 46.3 & 43.2 & 43.2 \\
  Arithmetic & 42.5 & 42.5 & 42.7 & 42.7 \\

    \bottomrule
    \end{tabular}
    \caption{Performance of \model~  for different answer types on \tatdqa~dev and test set. }
   
    \label{tab:model-types}
\end{table}

The \tatdqa~dataset offers four different answer types. To better reveal the effects of our model, we analyze the performance of the \model (\llmvt) on each of these answer types.
The results are summarized in \tabref{model-types}. 
From the table, we can see that the metric \fone{} on test set for the answer type of \emph{Span} shows the best results, while the results on \emph{Counting} and \emph{Spans} are similar, which is probably because these two types use similar prediction strategies in our model, i.e. generating the answer directly after obtaining the evidence using sequence tagging.
Comparably, the \emph{Arithmetic} type has the worst results of \fone{}, indicating that the discrete reasoning capability still demands further enhancement on complex real-world documents.

\subsubsection{Results of \model~model on \tatqa} 
To further test the effectiveness of our \model~ model, we also test it on the \tatqa~dataset.
In particular, we adapt the variant \model(\roberta) as there are only well-structured tables and relevant text and no document images in \tatqa.

Following \tagop, ~\model(\roberta)~takes the question, the flattened table by row and the associated paragraphs sequentially as input instead.
From \tabref{tatqa-metric}, we can observe that our \model~model significantly outperforms the state-of-the-art method on \tatqa.
It is worth mentioning that our model performs better on \tatqa~ compared with its performance on the new \tatdqa~ dataset.
This indicates that \tatdqa~ is more complex and challenging than \tatqa, which will be analyzed in detail at below.

\begin{table}[t]
    \centering
    \begin{tabular}{lrrrr}
    \toprule
     \multirow{2}{*}{\bf Method}    & \multicolumn{2}{c}{\bf Dev} & \multicolumn{2}{c}{\bf Test} \\
     \cmidrule(lr){2-3}
     \cmidrule(lr){4-5}
         & EM & \fone{} & EM&  \fone{}\\
    \midrule

    NumNet+ V2   &  38.1 & 48.3 & 37.0 & 46.9 \\
    \tagop   &  55.2 & 62.7 & 50.1 & 58.0  \\
    \addlinespace

    \model~(\roberta) & \bf 68.2 & \bf 76.8 & \bf 63.6 & \bf 72.7  \\
    % \uparrow &  +13.0\uparrow  & +16.1 \uparrow &  +13.5 \uparrow & +14.7\uparrow  \\

    \bottomrule
    \end{tabular}
    \caption{Performance of our model and baseline models on dev and test set of \tatqa. Best results are marked in bold.}
   
    \label{tab:tatqa-metric}
\end{table}

\subsubsection{The Challenges of \tatdqa}
\label{exp-challenge}
The new \tatdqa~ dataset differs from \tatqa~ in two aspects: 1) in \tatdqa~ there is no well-structured table and instead, all contents are organized with text blocks; 
2) the texts in \tatdqa~ are not necessarily associated to the table, which are with longer contents and more complex (as verified experimentally above).
Hereby, we try to reveal the challenges of \tatdqa~by answering following three questions:

\begin{itemize}
   \item  \textbf{Q1}: How do the tabular data in the document affect the model performance? 
   \item \textbf{Q2}: What is the difference in the model performance between multi-page documents and one-page documents?
  \item  \textbf{Q3}: What is the model performance on the documents with more content compared to those with less content?
\end{itemize}

\begin{figure}[t]
    \begin{center}
    \includegraphics[scale=0.3]{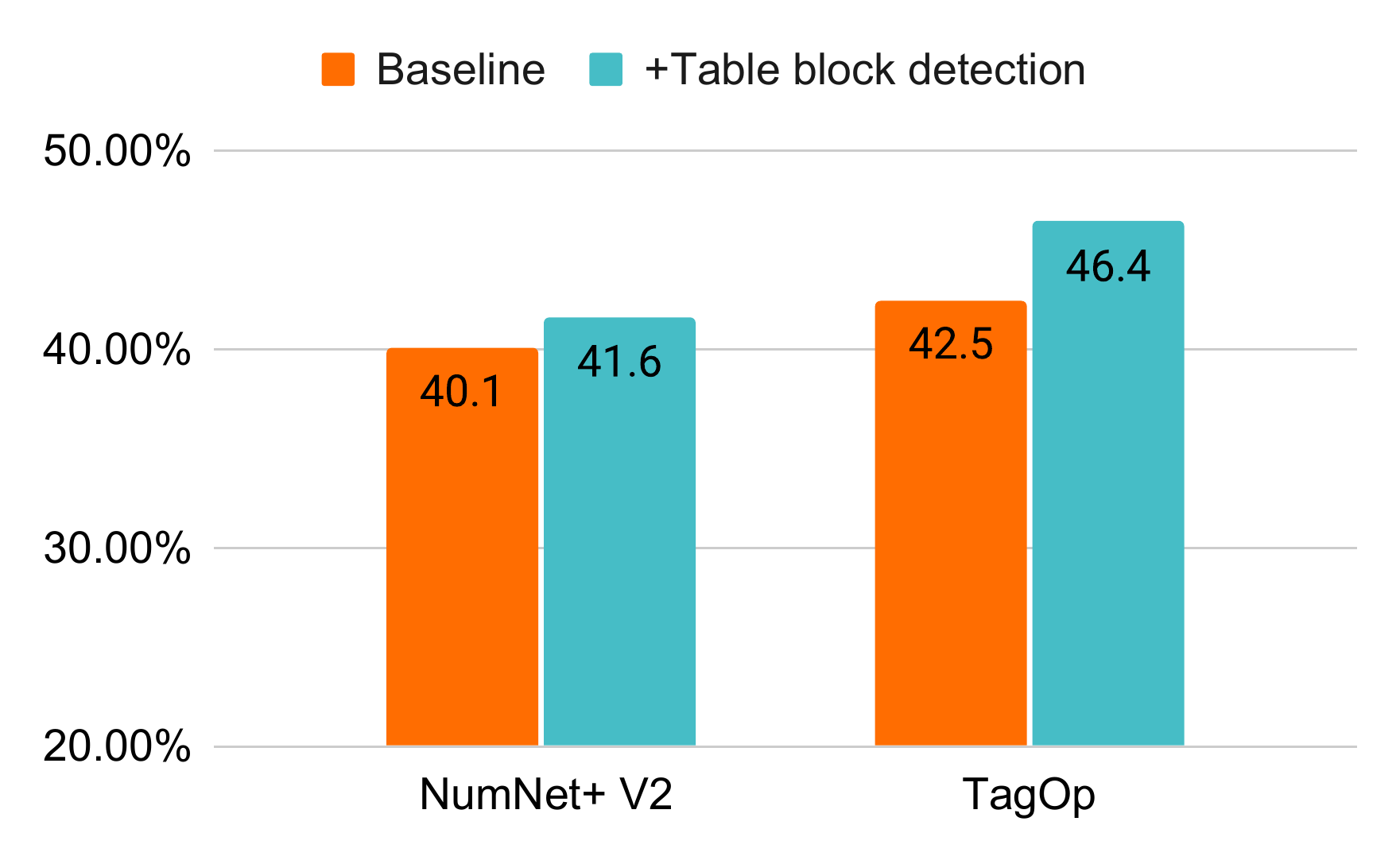}
    \end{center}
    \vspace{-1em}
    \caption{\label{fig:table-det} 
    The \fone{} score on \tatdqa~test set after we add the same table block detection module in NumNet+ V2 and \tagop~from \model.
   }
\end{figure}

For \textbf{Q1}, to evaluate the effect of tabular data in the document on the performance, we apply the same table block detection module in Section \ref{subsec4-1} to NumNet+ V2 and \tagop~models respectively.
The results are shown in \figref{table-det}.
We can observe that both NumNet+ V2 and \tagop~models gain better performance on  \tatdqa~test set after adding the table block detection module.
Particularly, the \tagop~model achieves $3.9\%$ higher \fone{} score than  original version, which is a significant increase.
This indicates that it is favorable to recognize tables in the document first and then process separately.

\begin{figure}[t]
    \begin{center}
    \includegraphics[scale=0.3]{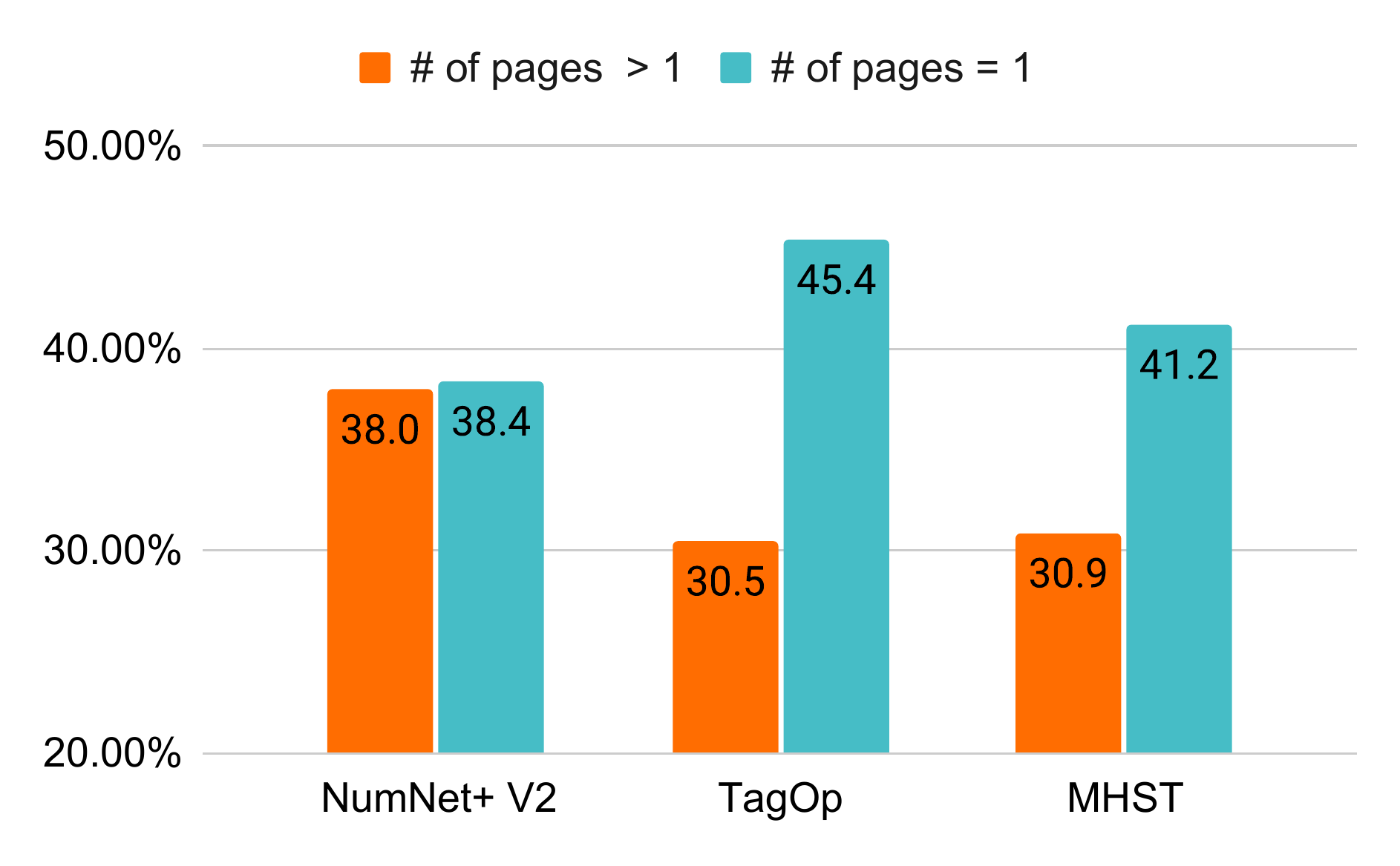}
    \end{center}
    \vspace{-1em}
    \caption{\label{fig:page-compare} 
    Performance comparison in \fone{} score between one-page documents and multi-page documents on  \tatdqa~ test set. Here \model~denotes the variant of \model (\roberta).
   }
\end{figure}

To answer \textbf{Q2}, we compare the performance of three models, NumNet+ V2, \tagop~ and \model~(\roberta), over multi-page documents and one-page documents, respectively.   
We here do not choose \model (\llmvt) as it takes as input the visual image, which is converted from the document page with the major table, meaning using only one page and hence unfair for comparison with other models.
\figref{page-compare} shows the comparison results on the test set of \tatdqa.
We can observe that the performance on multi-page documents is much worse than that on one-page documents for all three models.
Among them, \tagop~and \model(\roberta) perform significantly better on documents with only one page. 
The results indicate that multi-page documents are more challenging than one-page ones.

\begin{figure}[t]
    \begin{center}
    \includegraphics[scale=0.3]{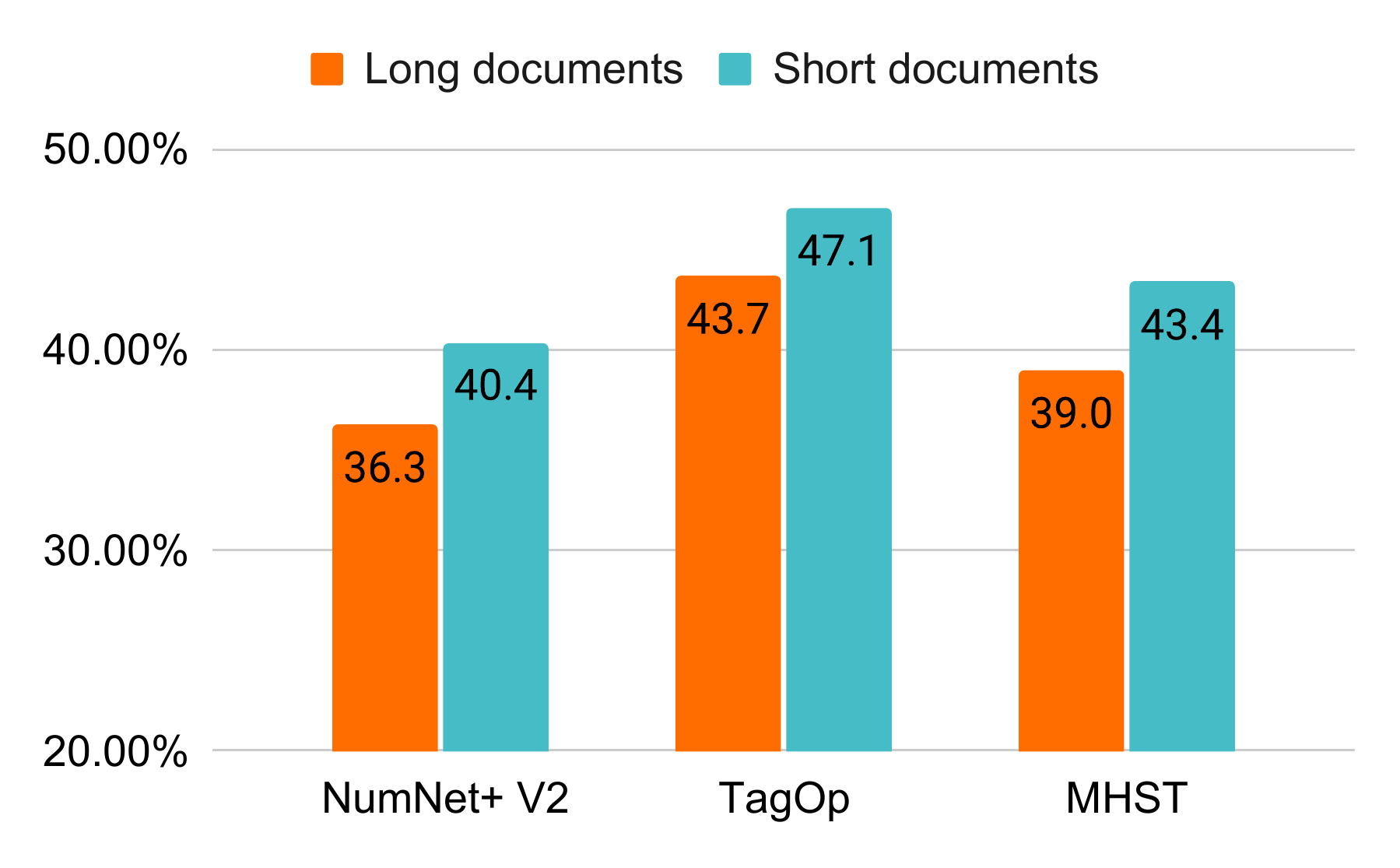}
    \end{center}
    \vspace{-1em}
    \caption{\label{fig:doc-len} 
    Performance comparison in \fone{} score between \texttt{Long} documents and \texttt{Short} document on \tatdqa~test set. Here \model~denotes the variant of \model~ (\roberta).
   }
\end{figure}

Then let us consider \textbf{Q3}. As shown in \tabref{tatdqa_stats}, the average number of OCR words per page is around 500 in \tatdqa, which is much larger than previous document VQA datasets DocVQA~\cite{Mathew2020DocVQA} and VisualMRC~\cite{VisualMRC2021}.  
We investigate the model performance with respect to the increase of document length for NumNet+ V2, \tagop~and \model(\roberta).  
For fairness, we only use the documents with one single page in the  \tatdqa ~test set for experiments here. 
We divide these one-page documents into two halves by the median length, with the half containing fewer words named as ``Short documents'' and the other containing more words as ``Long documents''.
The results are shown in \figref{doc-len}.
We can see that the performance on ``Short documents'' is significantly better than that on ``Long documents''.
Long documents are still a big challenge for document understanding tasks, where further research efforts are demanded.

\begin{table}[t]
\centering
\footnotesize
\vspace{-1em}
\begin{tabular}{p{0.1\linewidth}|p{0.14\linewidth}|p{0.63\linewidth}}
    \toprule
    \bf Module & \bf Error(\%) & \bf Example \\
    \toprule
    \multirow{6}{*}{\bf SAP} &  
    \multirow{3}{*}{\shortstack[l]{Offset \\ Error \\ (30\%)}}  
    & \textbf{Q}: How is the fair value of financial instruments that are not traded in active markets determined? \\
    & & \textbf{G}: \textcolor{blue}{using valuation techniques} \\
    & & \textbf{P}: \textcolor{red}{using valuation} \\
    \cmidrule{2-3}
    & \multirow{3}{*}{\shortstack[l]{No \\ Overlap \\ (7\%)}} 
    & \textbf{Q}: In 2018, why did the revenues grew across all regions? \\
    & & \textbf{G}:\textcolor{blue}{mainly due to growth in Imaging and Automotive.} \\
    & & \textbf{P}: \textcolor{red}{increase.} \\
    \midrule
    
    \multirow{6}{*}{\bf ETG} &  
    \multirow{3}{*}{\shortstack[l]{Wrong \\ Sign \\ (19\%)}}  
    & \textbf{Q}: What was the change in raw materials between 2018 and 2019? \\
    & & \textbf{G}: 36,987 - 45,333 = \textcolor{blue}{-8346} \\
    & & \textbf{P}: 45,333 - 36,987 = \textcolor{red}{8346} \\
    \cmidrule{2-3}
    & \multirow{3}{*}{\shortstack[l]{Wrong \\ Expressions \\ (15\%)}} 
    & \textbf{Q}: What is the percentage change in the short term investments between 2018 and 2019? \\
    & & \textbf{G}: \textcolor{blue}{(17,779 - 11,303)/11,303}\\
    & & \textbf{P}: \textcolor{red}{(11,303 - 17,779 )/17,779} \\
    \midrule
    
    \multirow{6}{*}{\bf ST} &  
    \multirow{3}{*}{\shortstack[l]{Wrong\\Taggings \\ (13\%)}}  
    & \textbf{Q}: What was the average Reductions for prior year tax positions from 2017-2019? \\
    & & \textbf{G}: ( -14 + \textcolor{blue}{0} + \textcolor{blue}{0})/3 \\
    & & \textbf{P}: (\textcolor{red}{2019} - 14 - 14)/3 \\
    \cmidrule{2-3}
    & \multirow{3}{*}{\shortstack[l]{Failed\\ Segment \\ (12\%)}} 
    & \textbf{Q}: What are the three operating and reportable segments? \\
    & & \textbf{G}: ``\textcolor{blue}{PSG}'', ``\textcolor{blue}{ASG}'', ``\textcolor{blue}{ISG}'' \\
    & & \textbf{P}: ``\textcolor{red}{PSG, ASG and ISG. The}''\\
    \midrule
    
     \multirow{3}{*}{\bf SP} &  
    \multirow{3}{*}{\shortstack[l]{Wrong \\ Scale \\ (4\%)}}  
    & \textbf{Q}: What was the difference between total Term Loans and total Future Lease Commitments? \\
    & & \textbf{G}:1,500 - 1,202 = 298 \textcolor{blue}{millions} \\
    & & \textbf{P}: 1,500 - 1,202 = 298 \textcolor{red}{thousands} \\
    \bottomrule
    \end{tabular} 
    \caption{
    Examples of errors and corresponding percentage in each module. 
    SAP, ETG, ST and SP are the abbreviations of Span Answer Predictor, Expression Tree Generator, Sequence Tagging Module and Scale Predictor respectively. 
    Q, G and P denote question, ground truth and prediction.}
\label{tab:error-case} 
\end{table}

\subsubsection{Error Analysis}
To further investigate our \model~model, we randomly sample $100$ error instances on the test set and analyze the reasons.
As shown in \tabref{error-case}, the errors occurred to four modules, Span Answer Predictor (SAP), Expression Tree Generator (ETG), Sequence Tagging (ST) module and Scale Predictor (SP), which are classified into seven categories (Col. 2 in \tabref{error-case}), each with an example.
We can observe that, 
1) \emph{SAP module} ($37$\%): $37$\% errors are due to inaccurate predictions of starting and ending positions for Span questions, i.e., $30$\% predictions overlapping but not exactly matching with ground truth, and $7$\% predictions having zero overlap with ground truth;
2) \emph{ETG module} ($34$\%): $34$\% of all errors are caused by ETG generating wrong expressions for the input of correct evidences from ST module, among which $19$\% are wrong number signs (i.e., positive/negative) and $15$\% are other wrong expressions;
3) \emph{ST module} ($25$\%): $25$\% errors are due to ST predicting wrong taggings, where interestingly, in $12$\% error cases, ST predicts a single string instead of identifying multiple answers from it for multi-span questions;
 4) \emph{SP module} ($4$\%): $4$\% errors are due to SP module failing to predict the scale of the answer.
We will further improve our model based on these error analysis findings.

\section{Conclusion and Future Works}

In this work, we propose a new challenging Document VQA dataset, named \tatdqa, constructed based on real-world high-quality financial reports, in which each document must have both tabular and textual data. 
With the new dataset, we further propose a novel \model~ model for addressing Document VQA tasks.
From our experiments, we have gained an insight that processing tabular data separately would significantly improve model performance.
This inspires us to explore the approaches of detecting and recognizing all tables in the given document and processing them differently to enhance the Document VQA models.
There are usually plentiful numbers in the document, which would overwhelm the capability of almost all existing discrete reasoning models.
One promising solution to such an issue is to differentiate the relevant numbers to the question from lots of numbers in the document before feeding them to the discrete reasoning module to derive the final answer, like what we have done in this work as an inspiring work.
In the future, we would like to continue exploring other potential methods to effectively identify the relevant numbers to a given  question to enhance the discrete reasoning over complex business documents.

\begin{acks}
The authors gratefully thank all the anonymous reviewers for their positive feedback.
This research is supported by the NExT Research Center, Singapore.
\end{acks}

\bibliographystyle{ACM-Reference-Format}
\bibliography{sample-sigconf}

% \appendix

\end{document}